%% file: main.tex
\documentclass[sigconf]{acmart}
\AtBeginDocument{%
  \providecommand\BibTeX{{%
    \normalfont B\kern-0.5em{\scshape i\kern-0.25em b}\kern-0.8em\TeX}}}

\usepackage{url}
\usepackage{soul}
\usepackage{mathrsfs}
\usepackage{hhline}
\usepackage{bm}
\usepackage{bbm} 
\usepackage{tabularx}
\usepackage{multirow}
\usepackage{graphicx} 
\usepackage{amsmath,amsfonts,amssymb}
\usepackage{wrapfig}
\usepackage{color}
\usepackage{booktabs}
\usepackage{setspace}
\usepackage{float}

\newcommand\BibTeX{B\textsc{ib}\TeX}
\newcommand{\nop}[1]{}

\copyrightyear{2021} 
\acmYear{2021} 
\setcopyright{acmcopyright}\acmConference[KDD '21]{Proceedings of the 27th ACM SIGKDD Conference on Knowledge Discovery and Data Mining}{August 14--18, 2021}{Virtual Event, Singapore}
\acmBooktitle{Proceedings of the 27th ACM SIGKDD Conference on Knowledge Discovery and Data Mining (KDD '21), August 14--18, 2021, Virtual Event, Singapore}
\acmPrice{15.00}
\acmDOI{10.1145/3447548.3467410}
\acmISBN{978-1-4503-8332-5/21/08}

\begin{document}
\fancyhead{}

\title{TopNet: Learning from Neural Topic Model to Generate\\ Long Stories}






\author{Yazheng Yang}\authornote{Equal contribution. $^\dag$ Work done while visiting the Ohio State University.}
\affiliation{%
  \institution{College of Computer Science}
  \institution{Zhejiang University}
  \city{Hangzhou}
  \country{China}}
  \email{yazheng\_yang@zju.edu.cn}
  
\author{Boyuan Pan$^\ast$$^\dag$}
\affiliation{%
  \institution{State Key Lab of CAD\&CG}
  \institution{Zhejiang University}
  \city{Hangzhou}
  \country{China}}
  \email{panby@zju.edu.cn}
  
\author{Deng Cai}
\affiliation{%
  \institution{State Key Lab of CAD\&CG}
  \institution{Alibaba-Zhejiang University Joint Institute of Frontier Technologies}
  \institution{Zhejiang University}
  \city{Hangzhou}
  \country{China}}
  \email{dengcai@cad.zju.edu.cn}
  
\author{Huan Sun}
\affiliation{%
  \institution{Department of Computer Science and Engineering}
  \institution{The Ohio State University}
  \city{Columbus}
  \country{USA}}
  \email{sun.397@osu.edu}

\input{abstract.tex}




\maketitle

\input{introduction.tex}

\input{model.tex}

\input{experiment.tex}

\input{related_work.tex}

\input{conclusion.tex}
\input{acks.tex}

\bibliographystyle{ACM-Reference-Format}
\bibliography{topnet}

\end{document}

%% file: abstract.tex
\begin{abstract}
	Long story generation (LSG) is one of the coveted goals in natural language processing. Different from most text generation tasks, LSG requires to output a long story of rich content based on a much shorter text input, and often suffers from information sparsity. In this paper, we propose \emph{TopNet} to alleviate this problem, by leveraging the recent advances in neural topic modeling to obtain high-quality skeleton words to complement the short input. In particular, instead of directly generating a story, we first learn to map the short text input to a low-dimensional topic distribution (which is pre-assigned by a topic model). Based on this latent topic distribution, we can use the reconstruction decoder of the topic model to sample a sequence of inter-related words as a skeleton for the story. Experiments on two benchmark datasets show that our proposed framework is highly effective in skeleton word selection and significantly outperforms the state-of-the-art models in both automatic evaluation and human evaluation.
\end{abstract}

%% file: introduction.tex
\section{Introduction}


{Long story} generation (LSG) is one of the desired goals for artificial intelligence systems and has many real-world applications such as automatic news generation, tutoring systems, \emph{etc}. Given a short text description or even a single word, the task of LSG is to teach the machine to generate a long narrative story (Table \ref{tab1}). The recent introduction of high capacity language models (\textit{e.g.,} GPT-2)~\cite{radford2019language} have shown their ability to generate stylistically coherent text but also suffer from uncontrollability in topics and content, which makes them have very few opportunities in industrial or commercial usage.


\begin{table}[t]
	\begin{center}
	\begin{spacing}{0.87}
		\begin{tabular}{p{0.9\columnwidth}}
			\toprule
			{\small \textbf{Short Description:} sunflower seeds}\\ \hline
			{\small \textbf{Story:} Misty took a bag of sunflower seeds downstairs without asking. This made her father irritated, but he    allowed her to do this. Then, she began spilling sunflower seeds. She spilled sunflower seeds even after her father said to be careful. Now, sunflower seeds are banned from the entire house.} \\ 
			\bottomrule
		\end{tabular}
		\end{spacing}
	\end{center}
	\caption{\label{tab1} An example from the ROCStories dataset.}
	\vspace{-0.7cm}
\end{table}

Most state-of-the-art works~\citep{fan2018hierarchical,xu2018skeleton,yao2019plan} tackle the challenges of LSG with a hierarchical structure, which first generates several skeleton keywords indicating the topic of the story, and then generates the story based on this skeleton. However, almost all of them generate the keywords by a supervised sequence-to-sequence model, which relies heavily on the maximum likelihood estimation objective and often leads to problems: sequences are dull, generic and repetitive~\citep{serban2016building,li2017adversarial}. Moreover, at the training stage, the labeled keywords are often created by word-frequency based methods or initialized by other sentence compression datasets, and suffer from lack of diversity and bias from different domains.\nop{rely heavily on interaction with humans or the keyword of each sentence in the story, which either limits generation at large scale or lacks enough details to support a long narrative story.}

Probabilistic topic modeling, which infers latent topics from documents is one of the greatest dimension reduction technologies. Given a corpus, the topic model assigns a topic distribution vector to each document and also builds a decoder that can reconstruct words in the document from the vector. This topic distribution contains the major information and latent features of the long document but is in the form of a low-dimensional vector, thus is an ideal distillation of the document and has been widely applied to various tasks such as text analysis and information retrieval~\citep{hofmann1999probabilistic,blei2003latent,teh2005sharing}. Recently, neural topic models have attracted much attention~\citep{kingma2014auto,gan2015scalable,miao2017discovering}. They typically approximate the posterior of a variational distribution with an inference model parameterized by a neural network, permitting unbiased and low-variance estimates of the gradients, and can provide a robust, scalable and theoretically sound foundation for long text modeling.

Inspired by the recent success of neural topic models, we focus on exploring whether they can further help provide knowledge for long story generation. Intuitively, the topic distribution assigned for each document by the topic model is a low-dimensional vector, which is easy for the given short description to map to; with the reconstruction decoder, it is also informative enough to generate diverse and inter-related topic words. Moreover, since the topic models are unsupervised learning methods, there is no need to annotate labels and they can be applied to large-scale datasets.

Given the above discussion, we propose \emph{TopNet}, a long story generation framework that leverages the neural variational inference~\cite{kingma2014auto} as in a topic model to tackle the information sparsity challenge in LSG (i.e., lack of information in the short input). Our \underline{key ideas} lie in two folds: (1) Given a short text input, predict the low-dimensional topic distribution of the to-be-generated story, rather than directly generate a word sequence. More specifically, we first compress each story in the training set into a topic distribution by the topic model and then train a \emph{Topic Generator} to map the short input text to the topic distribution. Since we employ the Gaussian distribution as the prior in the topic model, our variational topic distribution should also approach the same pattern that is easier for the input text to map to. Now, given a new input text, we can first use the Topic Generator to predict a topic distribution of its potential corresponding story and then use the reconstruction decoder of the topic model to transform the topic distribution to the word distribution, from which we can sample skeleton words to complement the short text. (2) Instead of randomly sampling from the decoded word distribution, we train a language model on the stories as an auto-regressive \emph{Word Sampler} to pick up more inter-related words as the skeleton. Finally, we concatenate these skeleton words with the given short text as input to Transformer~\citep{vaswani2017attention}, one of the state-of-the-art generative architectures, to generate a long story.

To verify the effectiveness of our approach, we conduct experiments on two long story generation tasks, {which differ in terms of the input text length and are representatives of two realistic application scenarios.}

(1) \emph{Title-to-article}. Given a title that usually consists of less than three words, the task requires to create a narrative article about it. On the ROCStories corpus~\citep{mostafazadeh2016corpus}, both automatic evaluation and human evaluation show that our TopNet significantly improves the performance over previous state-of-the-art approaches.

(2) \emph{Summary Expansion}. This task can be thought of the reverse of the text summarization task, and aims to expand the summary text by adding more details. We evaluate our model on the CNN/DailyMail~\citep{hermann2015teaching} dataset. The experimental results demonstrate that our generated stories perform much better in relevance, fluency and diversity than competitive baselines. Moreover, we show that our framework can not only augment the performance of normal size neural networks, but also has improvement on large-scale language models such as GPT-2.


%% file: model.tex
\section{Neural Topic Model}
In this section, we train a neural variational inference framework~\citep{kingma2014auto,miao2016neural,miao2017discovering} on the stories from the dataset with the goal of obtaining their latent topic distributions and a well-trained reconstruction decoder. In the next section, we will learn a map between the input text and the latent topic distribution and utilize the reconstruction decoder to sample informative keywords from the predicted topic distribution as a skeleton for long story generation.

\subsection{Parameterizing Topic Distributions}
Let $\bm{X} \in \mathbb{Z}^{D}_{+}$ denote the bag-of-words representation of a story, with ${Z}_{+}$ denoting nonnegative integers. $D$ is the vocabulary size, and each element of $\bm{X}$ reflects the frequency of the corresponding word in the story.
Following \citep{miao2017discovering}, we use a Gaussian random vector through a softmax function as the prior to parameterize the multinomial topic distribution. The generative process is:
\begin{equation}
\begin{aligned}
&\bm{t} \sim \mathcal{N}(\mu_{0}, \sigma_{0}^2), ~~ \theta = g(\bm{t}) \\ 
&z_n \sim {\rm Multi}(\theta), ~~ w_n \sim {\rm Multi}(\mathbf{\beta}_{z_n})
\end{aligned}
\end{equation}
where $\mathcal{N}(\mu_{0}, \sigma_{0}^2)$ is an isotropic Gaussian distribution, with mean $\mu_{0}$ and variance $\sigma_{0}^2$ in each dimension; $\mathbf{\theta} \in \mathbb{R}^{K}$ is the topic distribution of the story where $K$ is the number of topics; $z_n$ is the topic assignment for the observed word $w_n$; $g(\bm{t}) = {\rm softmax} (\mathbf{W}_g \bm{t} + \mathbf{b}_g)$, where $\mathbf{W}_g$ and $\mathbf{b}_g$ are trainable parameters; $\mathbf{\beta}_{z_n} \in \mathbb{R}^{D}$ represents the word distribution given topic assignment $z_n$. \nop{$\mathbf{\beta}_{z_n} \in \mathbb{R}^{D}$ represents the topic distribution over words given topic assignment $z_n$}

\begin{figure*}[t]
	
	\includegraphics[height=0.35 \textwidth]{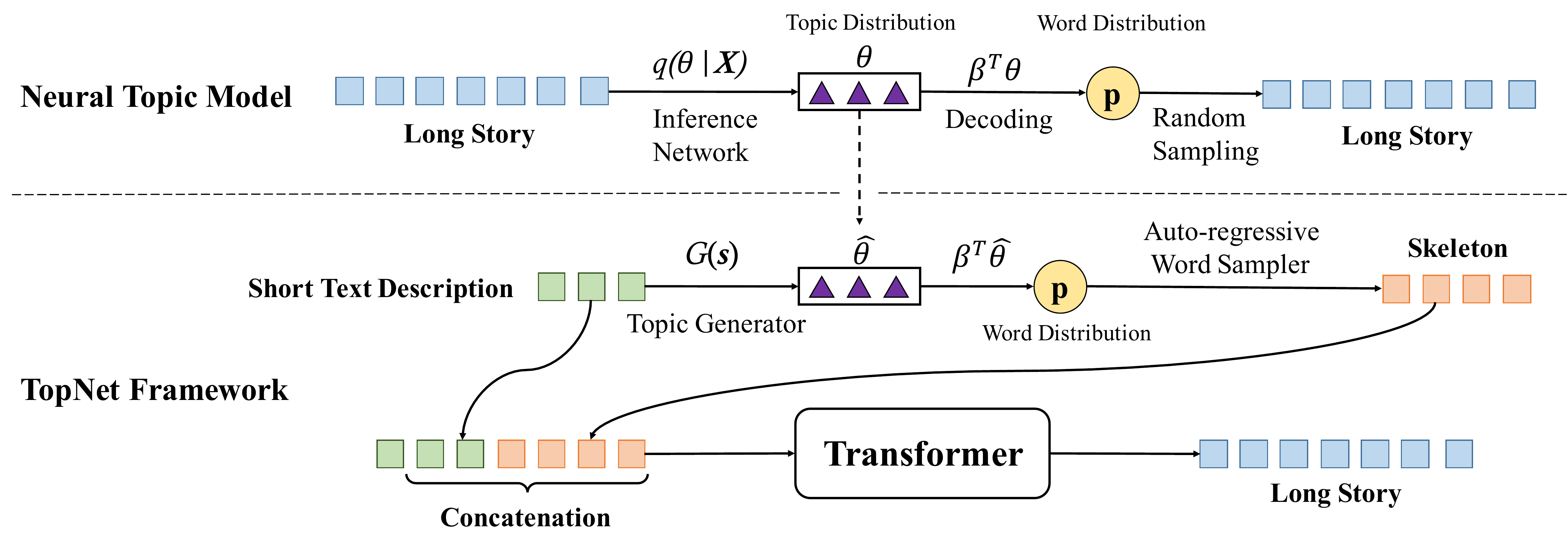}
	\caption{\label{fig1}Overview of our model, comprising the neural topic model (upper) and TopNet framework (bottom). \nop{The upper part is the neural topic model which is trained on the story collection, and the bottom part is the TopNet framework that predicts the topic distribution given a short text input.}}
\end{figure*}

\subsection{Neural Variational Inference}
\label{NVI-inf}
The neural variational inference is a simple instance of unsupervised learning where a continuous hidden variable $\theta$, which generates all the words in a document independently, is introduced to represent its semantic content. Inspired by~\citep{miao2017discovering}, we calculate the word distribution over topics by:
\begin{equation}
\begin{aligned}
\label{eq_v}
\mathbf{\beta}_{k} = {\rm softmax}(\mathbf{V} \cdot \mathbf{U}_{k}^{\top} )
\end{aligned}
\end{equation}
where $\mathbf{U} \in \mathbb{R}^{K \times H}$ is the trainable topic vectors, and $\mathbf{V} \in \mathbb{R}^{D \times H}$ is the word vectors. \nop{Different from previous works, we load the pre-trained word embeddings \emph{Glove}~\citep{pennington2014glove} as $\mathbf{V}$ and keep them fixed during training. This not only reduces the number of parameters, but also leverages the knowledge of word analogy that has been captured by pre-training on a large-scale text corpus.} Therefore, $\mathbf{\beta} \in \mathbb{R}^{K \times D}$ is the topic-to-word probability distribution matrix, which can be regarded as a decoder. The marginal likelihood for story $\bm{X}$ is:
\begin{equation}
\begin{aligned}
& p(\bm{X}|\mu_{0}, \sigma_{0}, \mathbf{\beta} ) \\
&= \int_{\mathbf{\theta}} p(\mathbf{\theta} | \mu_{0}, \sigma_{0}^2) \prod_{n} \sum_{z_n} p(w_n | \mathbf{\beta}_{z_n}) p(z_n | \mathbf{\theta}) d\mathbf{\theta}
\end{aligned}
\end{equation}

To parameterize the latent variable $\theta$, we construct a neural variational inference $q(\theta | \mu (\bm{X}), \sigma (\bm{X}))$ to approximate the posterior $p(\theta | \bm{X})$, where $\mu (\bm{X})$ and $\sigma (\bm{X})$ are functions of $\bm{X}$ that are implemented as multi-layer perceptrons (MLP). We optimize the variational objective function, also called the evidence lower bound (ELBO), as:
\begin{equation}
\begin{aligned}
\label{eq_loss}
\mathcal{L} = ~&\mathbb{E}_{q(\theta | \bm{X})} \Big [\sum_{n=1}^N {\rm log} \sum_{z_n} [p(w_n | \mathbf{\beta}_{z_n}) p(z_n | \mathbf{\theta})] \Big ] \\
&- D_{KL} [q(\theta | \bm{X}) || p(\mathbf{\theta} | \mu_{0}, \sigma_{0}^2)]
\end{aligned}
\end{equation}
where $q(\theta | \bm{X}) = q(\theta | \mu (\bm{X}), \sigma (\bm{X}))$. In practice, we re-parameterize $\theta = \mu (\bm{X}) + \epsilon \cdot \sigma (\bm{X})$ with the sample $\epsilon \in \mathcal{N}(0,I)$ to reduce the variance in stochastic estimation~\citep{kingma2014auto}. Since $p(\mathbf{\theta} | \mu_{0}, \sigma_{0}^2)$ is conditioned on a standard Gaussian prior, the KL term in Equation \ref{eq_loss} can be easily integrated as a Gaussian KL-divergence. Given a sampled $\theta$, the topic $z_n$ can be integrated out as:
\begin{equation}
\begin{aligned}
p(w_n|\beta, \theta) = \sum_{z_n} p(w_n | \mathbf{\beta}_{z_n}) p(z_n | \mathbf{\theta}) = \beta^{\top} \cdot \theta
\end{aligned}
\end{equation}
Hence we obtain a $K$-dimensional topic distribution $\theta$ for each story and a shared reconstruction decoder matrix $\beta$ where each row corresponds to one topic.

\section{TopNet Framework}
To generate a long story, one crucial challenge we have to attack is the lack of information on the input side (e.g., the length is much shorter than the target side). With little input information, the neural model degenerates to language model~\citep{fan2018hierarchical,devlin2018bert,radford2019language,joshi2020spanbert} that generates a story without taking the input into consideration. Such degeneration harms the fidelity of the generated story to its input as well as the story diversity. Intuitively, the above neural topic model captures what humans tend to write in a low-dimensional topic distribution $\theta$ and a knowledgeable reconstruction decoder $\beta$; {hence, given a short text, we will first map it to its potential topic distribution and then use $\beta$ to decode interesting skeleton words to complement the given short input, before generating a long story.}\nop{so given a short text we aim to map it to a reliable topic distribution instead of a long story sequence.}\nop{aim to approximate its potential topic distribution and use $\beta$ to decode it to interesting skeleton words.}\nop{and we aim to transfer the knowledge encoded in $\theta$'s and $\beta$ to the new story generation task.} 

Our proposed TopNet framework is shown in Figure \ref{fig1}. Its high-level idea is: Once a neural topic model is developed on a story corpus, we train a Topic Generator to map the short input text to a topic distribution, based on which, as well as the $\beta$ matrix, {we can map a new input text to a topic vector and then decode it to a word distribution to sample skeleton words for story generation.} 

\subsection{Topic Generator}
Given the short input text $\bm{s} = \{\bm{s}_1, \bm{s}_2, .., \bm{s}_m\}$ where $m$ is the number of the words, we use pre-trained word embeddings \emph{GloVe}~\citep{pennington2014glove} to transform the words into vectors $\{{\rm Emb}(\bm{s}_1), {\rm Emb}(\bm{s}_2),..., {\rm Emb}(\bm{s}_m) \}$. We use the average of these vectors, ${\rm Emb}_{mean}(\bm{s})$, to represent the input text and approximate its topic distribution via:
\begin{equation}
\begin{aligned}
\label{eq_mlp}
\hat{\theta} &= G(\bm{s})\\
&= {\rm softmax}(\mathbf{W}_1 \cdot {\rm ReLU}(\mathbf{W}_2 \cdot {\rm Emb}_{mean}(\bm{s}))
\end{aligned}
\end{equation}
where $\mathbf{W}_1, \mathbf{W}_2$ are trainable parameters. \nop{Do we actually need another model to estimate $\theta$? Can't we use the trained neural topic model to estimate the topic distribution of a new document?} For training, we use the short description text $\bm{s}$ from the training set and the mean of their topic distributions $\theta = \mu (\bm{X})$ computed by the topic model in Section~\ref{NVI-inf} to train the Topic Generator $G$, and we adopt the cross-entropy loss as our objective function. Note that given the low dimensionality of $\theta$ and the short input text, it is more practical to predict its topic distribution than, e.g., directly predict the word distribution of the to-be-generated story, as the latter is over the entire vocabulary. Moreover, the Gaussian prior of the variational distribution can also improve the robustness of the topic approximation.

\subsection{Auto-regressive Word Sampler}
\label{sec_lm}
Since $\beta$ is a shared topic-to-word matrix for all the stories, it stores the knowledge about the common constituents for various types of topics. We compute the word distribution $\mathbf{p}$ for a to-be-generated story by decoding its estimated topic distribution $\hat{\theta}$:
\begin{equation}
\begin{aligned}
\label{eq_p}
\mathbf{p} = {\rm softmax}(\beta^{\top} \cdot \hat{\theta})
\end{aligned}
\end{equation}
\nop{Hence we obtain a word distribution $\mathbf{p}$, indicates the probability of each word would appear in the story with the topic distribution $\hat{\theta}$.}

\nop{I think the flow gets disconnected here. Probably we can first describe your intuition for using a sequence generation model + top ranked words in p to generate complementary info, instead of directly selecting the top-K words in p? I feel this part is essential to sell the work and we should make it more appealing and convincing.} { To complement the short text input for long story generation, we aim to select skeleton words that can follow the content of the short input as well as interrelate with each other. Hence, instead of independently sampling a number of words from $\mathbf{p}$, we pre-train a forward language model on the collection of stories in the dataset as an auto-regressive \emph{Word Sampler}\nop{to select words that can follow the content of the short description as well as interrelate with each other}. Specifically, we use 3 layers of bi-directional Gated Recurrent Unit (GRU)~\citep{chung2014empirical} \nop{Do we need this 3-layer detail here? or move it to implementation details?} to form the language model and the top layer of the GRU output is used to predict the next token with a softmax function. We adopt the same vocabulary of the neural topic model in this language model and take out all the stop words in the stories. This means  our language model does not focus on syntactic structure, but aims to capture the intrinsic semantic coherence of a story.} 

{When sampling, we use the short description $\bm{s}$ as the initial input.} Formally, the language model computes the probability of a sampled word sequence by modeling the conditional probability: \nop{given the history:}
\begin{equation}
\begin{aligned}
&p(\bm{s}_1,\bm{s}_2, ... , \bm{s}_m, \bm{c}_1, \bm{c}_2, ..., \bm{c}_N)\\
&= \prod_{i=1}^{N}p(\bm{c}_{i}|\bm{s}_1,\bm{s}_2, ... , \bm{s}_m,  \bm{c}_1, \bm{c}_2, ..., \bm{c}_{i-1})
\end{aligned}
\end{equation}
where {$\bm{s_m}$ is the m-th word in the original short text input,} $\bm{c}_i$ is the sampled complementary word and $N$ is a fixed number.\nop{As the input at each time step, we combine the selected word at the previous step with the input text.}\nop{At each time step, we combine the selected word with previous input text as the new input.} We confine the language model to select words that are ranked in the top $N'$ ($N < N'$) of $\mathbf{p}$, by which we aim to obtain\nop{a skeleton that consists of a sequence of} the skeleton words $\{\bm{c}_1, \bm{c}_2, ..., \bm{c}_N \}$ that have a strong topical connection with the given short description and are also interrelated with each other so as to form a coherent\nop{narrative} story.\nop{which provides a proper range of choices.}  

\begin{table*}[h]

	\begin{center}
		\begin{tabular}{lccccc}
			\toprule
            Dataset & Training Set & Validation Set & Testing Set & Source Length & Target Length \\ \hline
			ROCStories  & 78529 & 9816 & 9816 & 2.2 & 52.0 \\
			CNN/DailyMail  & 287113 & 13368 & 11490 & 66.1 & 778.3 \\ 	
			\bottomrule
		\end{tabular}
	\end{center}
	\caption{\label{tab2} Statistics of the ROCStories and CNN/DailyMail datasets.} 
	\vspace{-0.48cm}
\end{table*}

\subsection{Story Generation}
To generate content-rich and coherent stories, we adopt the Transformer~\citep{vaswani2017attention} model which is known as good at drawing long dependencies and one of the state-of-the-art architectures for text generation. Transformer is a sequence transduction model based entirely on attention, replacing the recurrent layers most commonly used in encoder-decoder architectures with multi-headed self-attention. We concatenate the given short text description $\{\bm{s}_1, \bm{s}_2, .., \bm{s}_m\}$ with the skeleton $\{\bm{c}_1, \bm{c}_2, ..., \bm{c}_N \}$ as the input of the model, and the output is the story text. Note that our framework is not limited to the Transformer and other more advanced generative models can apply as well.

%% file: experiment.tex
\begin{table}[t]
	\begin{tabular}{lcc}
		\toprule
		Topic Model	& ROCStories & CNN/DM \\
		\midrule
		LDA & 617.13 & 1325.1\\ 
		NTM & \textbf{344.54} & \textbf{952.93}\\ 
		\bottomrule
	\end{tabular}
	\caption{\label{tab3}Perplexity of different topic models. ``NTM" denotes neural topic model. } 
	\vspace{-0.85cm}
\end{table}

\section{Experiments}
\subsection{Datasets}
We apply our approach to two long story generation tasks, which differ in terms of the input text length and are representatives of two realistic application scenarios, and conduct comprehensive analyses to show its effectiveness.

\noindent (1) \textbf{Title-to-article:} {Given a title that usually consists of less than three words, the task aims to create a narrative article about it.} In this task, we use ROCStories\footnote{\url{https://bitbucket.org/VioletPeng/language-model/src/master/Datasets/}}~\citep{mostafazadeh2016corpus}, which is a popularly used dataset whose input text is a short title and target is a five-sentence article, which captures a rich set of causal and temporal commonsense relations between daily events, making it a good resource for evaluating story generation models~\citep{peng2018towards,yao2019plan}. \\
(2) \textbf{Summary Expansion:} {This task can be thought of the reverse of the text summarization task, and aims to expand the summary text by adding more details.} We evaluate our model on CNN/DailyMail\footnote{\url{https://cs.nyu.edu/~kcho/DMQA/}}~\citep{hermann2015teaching} dataset, which is widely studied for text summarization tasks~\citep{see2017get,paulus2018deep}; in this paper we use it\nop{reverse the procedure} to evaluate models for expanding a short summary to a long story. 
Table \ref{tab2} shows the statistics of the datasets.

\subsection{Implementation Details}
\subsubsection{Neural Topic Model}
We train our neural topic model on the stories in a given dataset. The stories are preprocessed by stemming, filtering stopwords, and we choose the top 5000 most frequent words to compose vocabulary. We use \texttt{glove.840B.300d}~\citep{pennington2014glove} as the word vectors (Eq. \ref{eq_v}). For the inference network $q(\theta | \bm{X})$, we use an MLP with 2 layers and 500-dimension rectifier linear units. The dropout of 0.8 is applied to the output of the MLP before parameterizing the diagonal Gaussian distribution. The model is trained by Adam~\citep{kingma2014adam} and tuned by hold-out validation perplexity. We follow~\citep{miao2016neural} to alternately optimize the generative model and the inference network by fixing the parameters of one while updating the parameters of the other.

\subsubsection{TopNet Framework}
For the Topic Generator, we set the dimension of the weight matrix as 512. We optimize the model by Adamax~\citep{kingma2014adam} with a learning rate as 0.002. Our batch size is set as 128, and the dropout rate as 0.2. For the language model sampler, the dimension of the GRU is set as 512 and the model is trained with Adadelta~\citep{zeiler2012adadelta} with a learning rate of 0.001. The batch size is set as 20 and the dropout rate as 0.15. For the title-to-article task, we set the number of topics $K$ as 50, the $N$ and $N'$ in Section \ref{sec_lm} as 10 and 100 respectively and train the Transformer for 220k steps. For the summary expansion task, since its target length is relatively longer as shown in Table \ref{tab2}, we set the number of topics $K$ as 80, the $N$ and $N'$ in Section \ref{sec_lm} as 60 and 200 respectively and train the Transformer for 340k steps. For the Transformer model, we set hyperparameters the same as\nop{the settings in}~\citep{vaswani2017attention}, and use the base model in its official implementation\footnote{\url{https://github.com/tensorflow/models/tree/master/official/transformer}} for this work.

\begin{table*}[t]
	\begin{center}
		\begin{tabular}{l|cccc|c}
			\toprule
            Models & Inter-S & Intra-S & Dist-2 & Ent-4 & Dist-2 (SW) \\ \hline
			Inc-Seq2seq  & 0.95 & 0.16 & 0.074 &  7.929 & --\\
			Skeleton Model & 0.89 & 0.09 & 0.082 & 8.573 & 0.285 \\
			Static Planning &   0.82 & 0.06 & 0.093  & 9.238 & 0.473 \\ 
			Fusion Model & 0.71 & 0.05 & 0.101 & 11.558 & 0.604\\ \hline 	
			Transformer&  0.88 & 0.09 & 0.091 & 8.623 & --\\
			TopNet (LDA)  & 0.69 & 0.05 & 0.124 & 11.593 & 0.828 \\
			\textbf{TopNet} & \textbf{0.65} & \textbf{0.04} & \textbf{0.151}$^{~}$ & \textbf{11.682}$^{~}$ & \textbf{0.856}\\
			\bottomrule
		\end{tabular}
	\end{center}
	\caption{\label{tab4} Quantitative evaluation on the ROCStories dataset shows that our model achieves the state-of-the-art performance. ``Dist-2 (SW)" denotes the Dist-2 score for the generated skeleton words. \nop{The bottom part of the Table shows the performance of our TopNet and the ablation results.}} 
	\vspace{-0.48cm}
\end{table*}

\begin{table*}[t]

	\begin{center}
		\begin{tabular}{lcc|cc|cc}
			\toprule
		    \multicolumn{1}{l}{Choice \%} & \multicolumn{2}{c}{TopNet~ \emph{vs.} ~Fusion}~~~ & \multicolumn{2}{c}{TopNet~ \emph{vs.} ~Random~~} & \multicolumn{2}{c}{TopNet \emph{vs.} Top Words}  \\ \cline{2-7} \hline
			Coherence  & \textbf{55.1} & 44.9 & \textbf{73.5} & 26.5 & \textbf{56.4}  & 45.6 \\ 	
			Meaningfulness & \textbf{54.6} & 45.4 & \textbf{64.1} & 35.9 & \textbf{55.2} & 44.8\\
			Fidelity &  \textbf{52.0} & 48.0 & \textbf{78.3} & 21.7  & 47.5 & \textbf{52.5} \\
			Richness & \textbf{66.7} & 33.3 & \textbf{51.6} & 48.4 & \textbf{66.1} & 33.9\\
			\bottomrule
		\end{tabular}
	\end{center}
	\caption{\label{tab5} Human comparison with the state-of-the-art method and ablation study on the ROCStories dataset. } 
	\vspace{-0.48cm}
\end{table*}

\subsection{Baselines}
To demonstrate the effectiveness of our model, we choose several representative and state-of-the-art models from the current open-source works:\\
\textbf{Inc-Seq2seq}~\citep{bahdanau2015neural} denotes the incremental sentence-to-sentence generation baseline, which is built upon the Long-Short Time Memory Network (LSTM)~\citep{cheng2016long} along with attention mechanism. \\
\textbf{Skeleton Model}~\citep{xu2018skeleton} is also LSTM-based model, and trains its skeleton extraction module on other sentence compression datasets~\cite{filippova2013overcoming}. \\
\textbf{Fusion Model}~\citep{fan2018hierarchical} uses the human-annotated skeletons for training and adopts the gated self-attention along with Convolutional Sequence-to-Sequence Model~\citep{gehring2017convolutional} to learn the long-range context. \\
\textbf{Static Planning}~\citep{yao2019plan} uses word-frequency methods to train a skeleton extraction model and generates each sentence using each word of the skeleton.

\subsection{Topic Models}
Table \ref{tab3} presents the comparison of LDA and the neural topic model. As we can see, the neural topic model (NTM) performs much better than the traditional LDA on both datasets. The improvement over LDA indicates that the neural topic model can keep the information more accurately after compressing the original document, thus has potential to provide a better learning signal for the TopNet framework to generate a satisfying story.

\subsection{Title-to-article Task} In this task, we conduct experiments on the ROCStories dataset. For evaluation, we follow \citep{yao2019plan} to use \emph{inter-story} and \emph{intra-story} repetition scores, which denote the repetition rate of trigrams between stories at sentence level and the average trigrams repetition of sentences comparing with former sentences in a story respectively. We also follow~\citep{li2016diversity} to calculate \textit{Dist-2}, which is the proportion of unique bigrams over the total number of bigrams in the generated stories, and~\citep{zhang2018generating} to use the \textit{Ent-4} metric, which reflects how evenly distributed the 4-grams are over all generated stories. \nop{The smaller inter-story/intra-story scores and larger Dist-2/Ent-4, the more diverse the generated stories.} Since the input titles are usually less than three words, there can be various kinds of output that are all relevant so the evaluation metrics such as ROUGE~\cite{lin2004rouge} for generating a particular piece of text are not suitable for this task.

\paragraph{Results:} As shown in Table \ref{tab4}, our TopNet outperforms all the baseline models and achieves the state-of-the-art results, and the skeleton words generated by our method are more diverse than all other hierarchical models. 
In the bottom part, we conduct an ablation study: \nop{Firstly, we use the Transformer to generate the story only with the short title as the input. We can see that the performance drops a lot, which demonstrates that a short input is hard to support a much longer story without external knowledge.} We replace the neural topic model with LDA~\citep{blei2003latent}, and the performance is better than the Transformer's but still has a gap with the results of our TopNet. We conjecture that this is because the predicted top $N$ words for many stories are overlapped as we observe high-frequency words tend to be selected as the skeleton under the LDA model, while the neural inference network is more capable of learning complicated non-linear distributions.

For human evaluation in Table \ref{tab5}, we randomly sample 120 titles along with the generated stories and consider 4 aspects: \emph{Coherence} (whether the story as a whole is coherent in meaning and theme), \emph{Meaningfulness} (whether the story conveys some certain messages), \emph{Fidelity} (the relevance between the generated story and the title), \emph{Richness} (the amount of information in the story). We compare our TopNet with the Fusion Model~\citep{fan2018hierarchical}, Transformer with random words (randomly select words from the vocabulary as the skeleton), and Transformer with top $N$ words (directly select the top $N$ words from the probability $\mathbf{p}$ (Eq. \ref{eq_p}) instead of using our Word Sampler). 
For each sample, 5 people are asked to decide which of the two stories are better in the above aspects, and we show the average scores across the five annotators on all samples. As we can see, similar to the previous quantitative results, our TopNet almost outperforms their counterpart baselines in all evaluation aspects, thus demonstrating the effectiveness of the proposed framework.

\begin{figure}[t]
	\includegraphics[width=0.45 \textwidth]{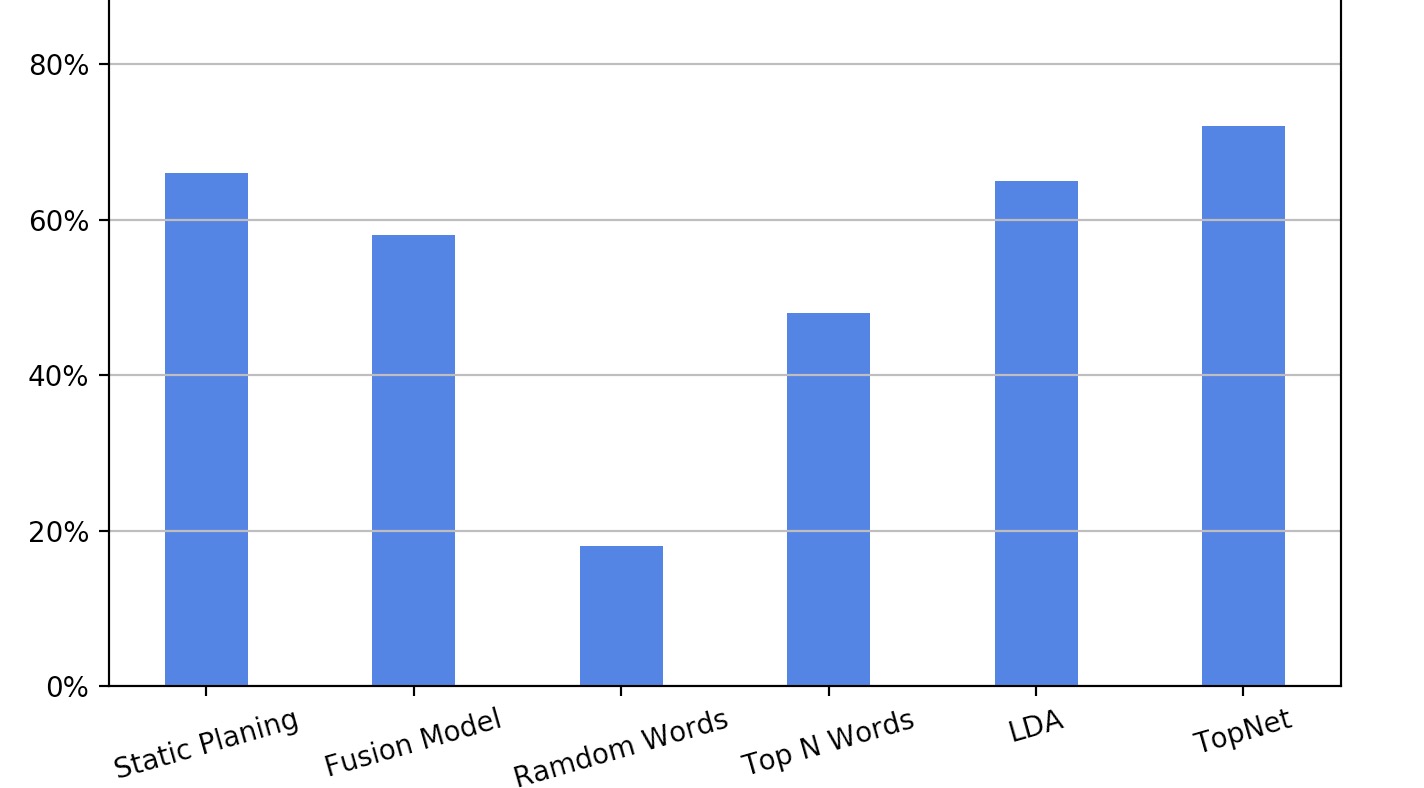}
	\caption{\label{fig2}Human accuracy at pairing the skeleton with the correct short input text (candidates consist of the original one and 4 randomly selected ones).}
	\vspace{-0.48cm}
\end{figure}

\begin{table}[t]
	\begin{center}
	\begin{spacing}{0.87}
		\begin{tabular}{p{0.95\columnwidth}}
			\toprule
			{\small \textbf{Title:} Allergies}\\
			{\small \textbf{Skeleton (Static Planning): }anna day doctor sick medicine}\\
			{\small \textbf{Skeleton (TopNet): }seasonal summer test medication sick symptom pill lunch need happy}\\ \hline
			{\small \textbf{Human: }Kia had a runny nose and headache for weeks. She finally went to her doctor. He told her she had allergies. He gave her an antihistamine to take. Soon Kia was feeling much better.}\\
			{\small \textbf{Static Planning: }Anna had a bad cough. She went to the doctor. The doctor told Anna she had a fever. Anna was very sick. She decided to go to the doctor.}\\ 
			{\small \textbf{TopNet: }Will had seasonal allergies. He sneezed and sniffled throughout the summer. He took allergy medication daily, which helped his symptoms. One morning he forgot to take a pill, and his allergies were terrible. Will made sure to take his pill with breakfast so he wouldn't forget.} \\ 
			\bottomrule
		\end{tabular}
		\end{spacing}
	\end{center}
	\caption{\label{tab6} Examples of skeletons and stories generated by human, Static Planning, and TopNet on the ROCStories dataset.}  
	\vspace{-0.6cm}
\end{table}

\begin{table*}[t]

	\begin{center}
		\begin{tabular}{l|cc|cc|c|c}
			\toprule
		    \multicolumn{1}{l}{\multirow{2}{*}{Models}} & \multicolumn{2}{c}{Relevance} & \multicolumn{2}{c}{Diversity} & \multicolumn{1}{c}{Fluency} & \multicolumn{1}{c}{Skeleton}   \\ \cline{2-7} 
		    \multicolumn{1}{c}{}  & RG-1  & RG-L    & Dist-2 & Ent-4   & Perplexity & Dist-2 \\ \hline
		    Inc-Seq2seq  & 22.4 & 13.8 & 0.051 & 10.024 & 38.73 & -- \\
			Fusion Model & 28.3 & 20.5 & 0.076 & 13.021 & 36.12 & 0.179\\ \hline
			Transformer & 25.1 & 16.9 & 0.060 & 11.565 & 38.15 & --\\
			Transformer + Random Words & 24.4 & 15.8 & 0.076 & 13.024 & 38.29 & \textbf{0.246}\\
			Transformer + Top $N$ Words & 27.9 & 19.4 & 0.070 & 12.754 & 36.38 & 0.127 \\
			TopNet (LDA) & 28.5 & 20.8 & 0.075 & 13.011 & 36.25 & 0.204\\
			\textbf{TopNet} & \textbf{29.3} & \textbf{21.6} & \textbf{0.078}$^{~}$ & \textbf{13.033}$^{~}$ & \textbf{35.73} & 0.219\\ \hline
			\hline
			GPT-2 & 11.2 & 8.3 & 0.121 & 14.380 & 30.25 & -- \\
			GPT-2 + TopNet & 12.8 & 9.6 & 0.114 & 14.133 & 30.21 & 0.219 \\
			\bottomrule
		\end{tabular}
	\end{center}
	\caption{\label{tab7} Quantitative evaluation on the CNN/DailyMail dataset, where the summary is the input and the article is the target. ``GPT-2 + TopNet" means we use the GPT-2 as the story generation module of the TopNet.}
	\vspace{-0.48cm}
\end{table*}

{In Figure \ref{fig2}, we show the correlation between the generated skeleton words and their short input text. We randomly pick up 100 $<$title, generated skeleton$>$ pairs and for each pair, we randomly sample 4 other titles from the dataset to mix them up with the original title. For each skeleton and the 5 titles, 5 people are asked to select the most relevant title according to the skeleton, and they obtain the highest averaged accuracy under TopNet's skeletons, which shows that our skeleton words are topical, meaningful, and loyal to the input text.}

In Table \ref{tab6}, we present a randomly selected title from ROCStories and the stories generated by humans (original story), Static Planning~\citep{yao2019plan}, and TopNet, and we also compare the skeletons generated by the latter two methods. It is obvious that our TopNet generates a more complicated and interesting story compared to the Static Planning and even the human-generated one. While the Static Planning produces redundant and repetitive content such as ``She decided to go to the doctor", our method generates a story that is much more informative and has more complex sentence structures. Since the number of the skeleton words in our TopNet is adjustable, we can produce more keywords than Static Planning, and thus provide more detailed information to support story generation. We also observe that some sampled words are not directly used in the generated story (\emph{e.g.} lunch, happy), but in general the complementary words are interrelated and pertinent to the given title, which can be attributed to\nop{benefit from} the latent feature extraction of the topic model. \nop{We show more examples in the supplementary material.}

\subsection{Summary Expansion Task} We further apply our model to another task named \textit{summary expansion}, which typically has a relatively longer input than the previous title-to-article task. In particular, we use the summaries in CNN/DailyMail dataset as the input to predict the corresponding stories. For evaluation, we measure the relevance of the generated story with the summary by ROUGE-1 and ROUGE-L since a summary is able to express some informative messages; we still use Dist-2 and Ent-4 to evaluate the diversity of the generated stories; we also measure the fluency of the stories by perplexity, which reflects how fluently the model can produce the correct next word given the preceding words. Except for ablation baselines, we select Inc-Seq2seq and Fusion Model as two representative baselines based on our previous experiments in Table \ref{tab4} and also compare our TopNet with GPT-2~\citep{radford2019language} (345M version)\footnote{\url{https://github.com/openai/gpt-2}}, which is a large-scale language model trained on external data and is known as one of the most powerful text generation models. \nop{I am actually looking for some brief explanation on why we use two different sets of baselines for the two tasks (Table 3 vs 5). Should we include some? Especially baselines are mentioned in 4.3, which seems to be for both tasks.}
\nop{I think you may briefly explain the differences between two tasks which lead to your choice of different models and different evaluation metrics.}

\paragraph{Results:} Table \ref{tab7} shows the performance of the models on the CNN/DailyMail dataset, and we can see that our TopNet achieves the best results over almost all the baselines, and our generated skeleton words also perform well in diversity comparing to other results (except random sampling). These results demonstrate that our TopNet is able to generate relevent, diverse yet fluent stories which further verifies the effectiveness of our approach. \nop{One interesting phenomenon is that the Transformer with random words performs worse under relevance metrics and influence than only using the summary as the input text. This is probably because when the complementary text is long enough, it can disturb the meaning of the original summary if the words are random and irrelevant}\nop{This may be explained by the length of the complementary text that is long enough to disturb the meaning of the original summary if the words are random and irrelevant}\nop{We also observe that unlike the metrics for diversity, the improvement of perplexity when adding random words to the Transformer is quite modest. This indicates that the random\nop{inappropriate} complementary input may not have a negative impact on the fluency of the generated text, but cannot substantially improve it either.}{For GPT-2, we see that the results of diversity and fluency are much better than the other approaches since it is pre-trained on an extremely large corpus and the model itself is much more complex\nop{much larger} than the others. However, we observe the articles it generates are almost irrelevant to the original ones, and many of them even deviate from the topic of the summaries. This indicates that although the large-scale language models have impressive improvements under various metrics\nop{over previous methods}, it is still challenging to control their generated text content, which is to some extent improved by our TopNet according to the results under relevance.\nop{how do we see this? by GPT-2 + TopNet? Reviewers might challenge this part by asking (1) do you have examples to show these statements? (2) why is there no human evaluation for this task?} } Such phenomenon also reflects that neural model is easy to ignore the model input when it contains little information for the model to take into consideration. With the informative supplement, our TopNet can theoretically remedy this problem.

Table \ref{tab_cnn} present an example of the generated skeleton and stories generated by human, GPT-2, and TopNet on the CNN/DailyMail dataset. 
As we can see, most of our generated skeleton words have strong correlation with the given summary, such as ``power", ``energy", ``facility", ``collapse", \emph{etc}. These words together with the summary support a long and coherent story. However, there are some words like ``lab", ``foot" that seem to be less relevant to the given summary, thus they don't play significant roles in story generation. Moreover, some words, such as ``Beijing", induces the model to generate the sentence ``The plant is located in the nearby city of Beijing, about 240 kilometers (150 miles) north of tokyo", but it doesn't make sense because it is a wrong information.

As for the generated stories, both our method (TopNet) and GPT-2 have produced a long and interesting story. We notice that the story generated by GPT-2 is longer and has more complex choices of words, and the GPT-2 is very good at making up specific name entities such as address, date, name and organization. However, we observe that the story generated by GPT-2 has deviated from the given summary, let alone the original story in the CNN/DailyMail dataset. It seems the GPT-2 is more skillful in continue writing, rather than expanding the given short text by adding details. We also find that the story generated by our TopNet has repetitive phrases such as ``since the march 11 earthquake and tsunami", indicating that the regular generators (\emph{e.g.} Transformer) still have a lot of room for improvement on content selection, even if provided with sufficient knowledge at the input side.

\begin{table*}[t]
	\begin{center}
		\begin{spacing}{0.87}
			\begin{tabular}{p{1.95\columnwidth}}
				\toprule
				{\small \textbf{Summary:} The operator of the fukushima nuclear plant said it has abandoned a robotic probe inside one of the damaged reactors. A report stated that a fallen object has left the robot stranded. The robot collected data on radiation levels and investigated the spread of debris.}\\
				{\small \textbf{Skeleton: }possible power energy inch facility core light announce improve identify tsunami lab expert quake unit safe confirm meter strike range emergency effort foot probe problem atmosphere electronic damage beijing image locate screen earth produce battery investigate fuel firm worker collapse staff source billion statement operate zone contribute conference since recover surface incident locate storm human heat current weather increase mile}\\ \hline
				{\small \textbf{Human: }The operator of the crippled Fukushima daiichi nuclear plant has given up trying to recover a robotic probe after it stopped moving inside one of the reactors. Tokyo electric power company deployed the remote-controlled robot on Friday inside one of the damaged reactors that had suffered a meltdown following a devastating earthquake and tsunami in 2011. It was the first time the probe had been used. The robot, set out to collect data on radiation levels and investigate the spread of debris, stalled after moving about 10 meters, according to a statement released by tepco. A newly released report and footage from the robot shows that a fallen object had blocked its path and left it stranded. Tepco decided to cut off the cable connected to the device sunday as it had already covered two-thirds of the originally planned route. It managed to collect data on radiation levels in 14 of the 18 targeted locations. Four years after the devastating nuclear crisis, the radiation levels inside the three damaged reactors are still extremely high and remain unsafe for people to enter. Decommissioning work is estimated to cost \$50 billion and will take years to complete. Tepco called the robotic probe an ``unprecedented'' experiment. CNN's Yoko Wakatsuki reported from Tokyo, Japan and Maomi Ng wrote from Hong Kong.}\\
				{\small \textbf{GPT-2: }A senior researcher at the Medecins Sans Frontieres study said southern regions of the Fukushima Daiichi nuclear power plant had past radiation warnings from Japanese health authorities. "The elderly had two 30mm fukushima radiators installed on top of them. In October last year the Vienna air and ground report Socio-Economic Department adopted most comprehensive rules on how the elderly should measure radiation levels if they use the radiators." In Japan technicians received an experimental Japanese Shishii add-on for Onsen fire bomblets proclaiming King Abe will rise up with 256 Great Komeito Attackers The Japan Meteorological Agency released this special report accuses Japan of having chemical weapons in spring 1986. Many Inland Lobsters were loaded with Anbak bombs in "standard package". a report described the outbreak as a "heavy blow, which operated simultaneously. A barrage of weapons and chemicals inside the nineties generation should have been spent. There was no evidence of evidence of contaminated equipment here. Everyone seems to talk about this as evidence of old bombs. In Texas, when Nichols received a shipment of \$40,000, he started recording the airport log of security personnel for weightlifting authorities. every flight time is recorded; most of the time, the official touch-screen window is detached from the flight HUD. Richard Yates interviews employees of Honda, Kentucky steel company, about retired attack helicopters being "rendered "ignorant", and that, after contracting Elvira for combat helicopter for close-up spraying in 2002, ...}\\ 
				{\small \textbf{TopNet: }The japanese operator of the nuclear power plant said it has abandoned a robotic probe inside one of the damaged reactors at the Fukushima daiichi nuclear power plant. The operator of the plant said it has abandoned the power plant since the march 11 earthquake and tsunami that resulted in the worst nuclear accident since Chernobyl, as a result of the heavy pressure on the plant 's surface. ``This is not the first time we have seen from outside the plant," said Tom Biden, director of the plant's mission, in a written statement. ``The nuclear power plant has not had any such problems with the power plant, as the cores of reactors at the Fukushima daiichi nuclear plant overheated and spewed radioactive material into the atmosphere." The plant is located in the nearby city of Beijing, about 240 kilometers (150 miles) north of tokyo. ``It has since been called off," the company said, "Adding that it does not have any reports of radiation contamination from the plant." The plant has been operating since the March 11 earthquake and tsunami, according to the U.S. environmental protection agency. The agency said in a statement that the robot collected data on radiation levels and investigated the spread of the reactors. "we are working closely with the engineers to determine the cause of the accident, "the agency said.} \\ 
				\bottomrule
			\end{tabular}
		\end{spacing}
	\end{center}
	\caption{\label{tab_cnn} Examples of the generated skeleton and stories generated by human, GPT-2, and TopNet on the CNN/DailyMail dataset.}  
	\vspace{-0.55cm}
\end{table*}

%% file: related_work.tex
\section{Related Work}
\subsection{Long Story Generation}
\nop{Generally speaking, long story generation aims to generate a long narrative story from any given input, such as short text descriptions~\citep{xu2018skeleton,yao2019plan}, long context~\citep{liu2018generating}, or even visual images~\citep{huang2016visual}, \emph{etc}. Considering a strong and general AI system should perform language generation at similar levels of creativity and originality as humans, we are particularly interested in generating long stories from the \textit{short} input text, rather than situations where the semantics of the target story are already fully specified by a long and informative input.\nop{get rid of situations where the semantics of the target are fully specified by the source and focus on generating long stories from much shorter inputs} }

\nop{Prior research for long story generation mainly relied on template-based approaches or stayed at the level of story plot planning without surface realization~\citep{porteous2009controlling,riedl2010narrative,li2013story}.} 

Recently, deep learning models have been demonstrated effective for LSG~\citep{kiros2015skip,roemmele2016writing,jain2017story,fan2019strategies,tambwekar2019controllable,goldfarb2019plan}. Most state-of-the-art methods~\citep{fan2018hierarchical,xu2018skeleton,yao2019plan} proposed to decompose the story generation procedures with a hierarchical generation strategy to first produce a skeleton and then generate a story based on the skeleton.
However, \nop{the skeletons in these work  are not}{these work either do not have skeletons} informative enough to support a long story, or they require human annotators to label skeletons for training. {Unlike previous work, we integrate the tricks of topic modeling into the story generation task through projecting an input text to its latent topic space and leveraging the reconstruction decoder of a neural topic model to obtain abundant\nop{adequate} inter-related skeleton words {in an unsupervised fashion} for story generation.}

{Large-scale language models (\emph{e.g.}, GPT-2) have also achieved impressive performance on long text generation~\citep{radford2019language,mao2019improving}. However, training such a language model can cost in excess of \$10,000 and also requires a huge amount of external data~\citep{wang2019language}, making it prohibitively expensive and time-consuming to perform a fully-fledged model exploration. Moreover, large-scale language models are usually uncontrollable because the local dependencies among the story itself are easier to model than the subtle dependencies between the input text and the story~\citep{fan2018hierarchical}. 
Different from tasks such as text abstract summarization~\citep{see2017get} or Neural Machine Translation (NMT)~\citep{gu2016incorporating}, where the semantics of the target are fully specified by the source, the generation of stories from the model input is far more open-ended. How to tackle the degeneration problem of story generation model that the neural models ignore the input and focus purely on previously generated sequence, is crucial in the scenario of generating long story. In such situation, it is hard for GPT-2 to produce loyal stories.
In this paper, we make full use of the given corpus by unsupervised learning while eliminating the need of large-scale external data or human interactions, and show that generating an informative skeleton for LSG is still important and challenging.}

\nop{Hence, efficiently generating a series of coherent and informative topics that can support the story is by far one of the most challenging and crucial problems for long story generation.}

\subsection{Topic Models}
Topic models have been studied for a variety of applications in document modeling and information retrieval. Beyond LDA~\citep{blei2003latent} that consists of multiple layers of Bayesian networks, various extensions have been explored to discover topics~\citep{teh2005sharing}, model temporal dependencies~\citep{blei2006dynamic}, among many others. Recently, neural topic models have gained much attention~\citep{hinton2009replicated,larochelle2012neural,gan2015scalable,miao2016neural,miao2017discovering}, which improved the performance of traditional methods. 
Different from prior approaches that sought closed-form derivations on the discrete text to train the topic model, \citet{miao2016neural} presented a generic variational inference framework for the intractable distributions over latent variables. Given the discrete text, they build an inference network in order to obtain the variational distribution. They further proposed to provide parameterisable distributions over topics~\citep{miao2017discovering}, so that it will permit training with back-propagation in the framework of neural variational inference, which makes it easy to obtain powerful topic models. 
{In this work, we also construct our topic model by a variational inference framework \cite{miao2016neural,miao2017discovering}, in which the latent topic vector is constrained by a prior distribution (\emph{e.g.} Gaussian distribution) and is easier to be approximated from the input text.}

\subsection{Topic Learning in NLP}
The idea of using learned topics to improve NLP tasks has been explored previously, including methods combining topic and neural language models~\citep{mikolov2012context,ahn2016neural,dieng2016topicrnn,lau2017topically,wang2018topic}, leveraging topic and word embeddings~\citep{liu2015topical,xu2018distilled}, as well as topic-guided VAEs~\citep{wang2019topic}. 
\citet{wang2018topic} presented a Topic Compositional Neural Language Model (TCNLM), which can not only maintain the overall topic representing the global information, but also learn the local semantic information of a document. 
In order to discriminate the ubiquitous homonymy and polysemy, \citet{liu2015topical} utilized the topic model to assign topic vector to each word embedding, and finally obtained the topical word embeddings. \citet{xu2018distilled} proposed to conduct the joint training of the word embedding and topic modeling to obtain topic-aware word embedding.
 \citet{wang2019topic} extended the VAE through specifying the prior of Gaussian distribution parametrized by latent topic distribution.
{Unlike previous works that mainly focus on modeling the latent topics, we propose to leverage the topic-to-word decoder of a topic model as well, which is knowledgeable and robust to help obtain skeleton words based on the approximated latent topics.} 
\vspace{-0.091cm}

%% file: conclusion.tex
\section{Conclusion}
In this paper, we propose a novel framework \emph{TopNet}, which leverages the \nop{advances}document modeling of the neural topic model for long story generation. To tackle the information sparsity\nop{knowledge deficiency} challenge, we learn to map the short input text to a low dimensional topic distribution and use the decoder of the topic model to transform it to the word distribution, based on which we finally develop a language model to sample skeleton words for story generation.\nop{we train neural networks to map the short input text to a low dimensional topic distribution, and then use the decoder of the topic model to transform it to the word distribution and finally employ a language model to sample complementary words for the story generation model.} The experimental results on different tasks show that our model significantly outperforms the state-of-the-art approaches and can produce interesting and coherent stories. We also demonstrate that our method has the potential to improve the large-scale language model by providing appropriate topic words. Future works include further research into controllable long sequence generation as well as exploring unsupervised or self-supervised learning in other NLP tasks. \nop{as well as exploring other scenarios for generation such as dialog and videos. and the future works involve other generation tasks such as dialog, videos, etc.}

%% file: acks.tex
\begin{acks}
This work was supported in part by The National Key Research and Development Program of China (Grant Nos: 2018AAA0101400), The National Nature Science Foundation of China (Grant Nos: 62036009, 61936006), the Alibaba-Zhejiang University Joint Institute of Frontier Technologies, and The Innovation Capability Support Program of Shaanxi (Program No. 2021TD-05). Boyuan Pan (while visiting the Ohio State University) and Huan Sun were sponsored in part by the Army Research Office under cooperative agreements W911NF-17-1-0412, NSF Grant IIS1815674,
NSF CAREER \#1942980, Fujitsu gift grant, and Ohio Supercomputer Center \cite{OhioSupercomputerCenter1987}. The views and conclusions contained herein are those of the authors and should not be interpreted as representing the official policies, either expressed or implied, of the Army Research Office or the U.S. Government. The U.S. Government is authorized to reproduce and distribute reprints for Government purposes notwithstanding any copyright notice herein.
\end{acks}
\vspace{-2.5pt}